\documentclass{aamas2017}
\usepackage{hyperref}

\usepackage[latin1]{inputenc}
\usepackage{graphicx}
\usepackage{subcaption}
\usepackage{amsmath}
\usepackage{dsfont}
\usepackage[capitalize]{cleveref}
\usepackage{amssymb}
\usepackage{mathtools}

\usepackage{amsthm}
\theoremstyle{plain}

\newtheorem{MyProposition}{Proposition}

\newtheorem*{myRemark}{Remark}

\newcommand{\mat}[1]{\mathrm{\mathbf{#1}}}
\newcommand{\indicator}{\mathds{1}}

\newcommand{\E}{\mathbb{E}}
\newcommand{\Eof}[1]{\E\left[#1\right]}
\newcommand{\EofP}[2]{\E_{#1}\left[#2\right]}

\DeclareMathOperator*{\argmax}{arg\,max}

\newcommand{\prob}{P}

\newcommand{\given}{\mid}

\newcommand{\trans}{^{\mathsf T}}

\newcommand{\eqpunkt}{.} 
\newcommand{\eqkomma}{,}
\newcommand{\defeq}{\coloneqq}

\newcommand{\ie}{i.e.}
\newcommand{\eg}{e.g.}

\usepackage{tikz}
\usetikzlibrary{positioning,fit,shapes,backgrounds}
\tikzstyle{vertex}=[circle, line width=1.5pt, draw=black!100, fill=black!10]
\tikzstyle{template}=[rectangle, line width=1pt, inner sep = 0.5cm, draw=black!100, rounded corners=3mm]
\tikzstyle{block}=[rectangle, line width=1pt, draw=black!100, fill=black!10]
\tikzstyle{action}=[circle, line width=1.25pt, draw=black!100, fill=black!10]
\tikzstyle{reward}=[circle, line width=1.25pt, draw=black!100, fill=black!10]
\tikzstyle{agent}=[circle, line width=1.25pt, draw=black!100, fill=black!10]
\tikzstyle{swarm}=[circle, line width=1.25pt, draw=black!100, fill=black!10]
\tikzstyle{background}=[rectangle, draw=black, fill=white, rounded corners=10mm]
\tikzstyle{connect}=[rectangle, draw=black!20, fill=black!20, rounded corners=2mm]
\tikzstyle{policy}=[line width=1.5pt, draw=black!100, fill=black!10]
\tikzstyle{objective}=[line width=1.5pt, draw=black!100, fill=black!10]
\tikzstyle{state}=[line width=1.5pt, draw=black!100, fill=black!10]

\def \sizeA {5.3cm}
\def \sizeB {5.1cm}

\begin{document}

\title{Inverse Reinforcement Learning in Swarm Systems}

\numberofauthors{1}

\author{
	\alignauthor
	Adrian \v{S}o\v{s}i\'{c}, Wasiur R. KhudaBukhsh, Abdelhak M. Zoubir and Heinz Koeppl \\[0.5cm] Department of Electrical Engineering and Information Technology\\ Technische Universit\"at Darmstadt, Germany 
}

\maketitle

\begin{abstract}
Inverse reinforcement learning (IRL) has become a useful tool for learning behavioral models from demonstration data. However, IRL remains mostly unexplored for multi-agent systems. In this paper, we show how the principle of IRL can be extended to homogeneous large-scale problems, inspired by the collective swarming behavior of natural systems. In particular, we make the following contributions to the field: 1)~We introduce the swarMDP framework, a sub-class of decentralized partially observable Markov decision processes endowed with a swarm characterization. 2)~Exploiting the inherent homogeneity of this framework, we reduce the resulting multi-agent IRL problem to a single-agent one by proving that the agent-specific value functions in this model coincide. 3)~To solve the corresponding control problem, we propose a novel heterogeneous learning scheme that is particularly tailored to the swarm setting. Results on two example systems demonstrate that our framework is able to produce meaningful local reward models from which we can replicate the observed global system dynamics.
\end{abstract}


\section{Introduction}
\noindent Emergence and the ability of self-organization are fascinating characteristics of natural systems with interacting agents. Without a central controller, these systems are inherently robust to failure while, at the same time, 
they show remarkable large-scale dynamics that allow for fast adaptation to changing environments \cite{buhl2006disorder,couzin2009collective}. Interestingly, for large system sizes, it is often not the complexity of the individual agent, %
but the (local) coupling of the agents 
that predominantly gears the final system dynamics.
It has been shown \cite{omel2008chimera,vicsek2012collective}, in fact, that even relatively simple local dynamics can result in various kinds of higher-order complexity at a global scale when coupled through %
a network with many agents. %
Unfortunately, the complex relationship between the global behavior of a system and its local implementation %
at the agent level %
is not well understood. In particular, it remains unclear when \mbox{-- and} %
how  -- a global system objective can be encoded in terms of local rules, and what are the requirements on the complexity of the individual agent in order for the collective to fulfill a certain task. %
Yet, this understanding is key to many of today's and future applications, such as distributed sensor networks \cite{lesser2012distributed}, nanomedicine~\cite{freitas2005current}, %
programmable matter \cite{goldstein2005programmable}, and self-assembly systems \cite{whitesides2002self}. %

A promising concept to %
fill this missing link is inverse reinforcement learning (IRL), which provides a data-driven framework for learning behavioral models from expert systems \cite{zhifei2012}. In the past, IRL has been applied successfully in many disciplines and the learned models were reported to even outperform the expert system in several cases \cite{abbeel2010autonomous,michie1990cognitive,sammut1992learning}. Unfortunately, IRL is mostly unexplored for multi-agent systems; %
in fact, %
there exist only %
few models which transfer the concept of IRL to systems with more than one agent.  %
One such example is the work presented in~\cite{natarajan2010multi}, where the authors extended the IRL %
principle to non-cooperative multi-agent problems in order 
to learn a joint reward model %
that %
is able to explain the system behavior
at a global scale. %
However, the authors assume that all agents in the network are controlled by a central mediator, an assumption which is clearly 
inappropriate for self-organizing systems. A decentralized solution was later presented in \cite{reddy2012inverse} but the proposed algorithm is based on the %
simplifying assumption that all agents are %
informed about the global state of the system. %
Finally, the authors of \cite{dufton2009multiagent} presented a multi-agent framework based on mechanism design, which can be used 
to refine a given reward model in order to promote a certain system behavior. However, the framework is not %
able to learn the %
reward structure entirely from demonstration data.

In contrast to previous work on multi-agent IRL, we do \textit{not} aspire to find 
a general solution for the entire class of multi-agent systems; instead, we focus on the important sub-class of homogeneous systems or \textit{swarms}. Motivated by the above-mentioned questions, we present a scalable IRL solution for the swarm setting to learn a single \textit{local} reward function which explains the \textit{global} behavior of a swarm, and which can be used to reconstruct this behavior from local interactions at the agent level. 
In particular, we make the following contributions: 1)~We introduce the \mbox{\textit{swarMDP}}, a formal framework to compactly describe homogeneous multi-agent control problems. 2)~Exploiting the inherent homogeneity of this framework, we show that the resulting IRL problem can be effectively reduced to the single-agent case. %
3)~To solve the corresponding control problem, we propose a novel heterogeneous learning scheme that is particularly tailored to the swarm setting. %
We evaluate our framework on two well-known system models: the Ising model and the Vicsek model of self-propelled particles. The results demonstrate that our framework is able to produce meaningful reward models from which we can learn local controllers that 
replicate the observed global system %
dynamics.

\section{The swarMDP Model}
\label{section:TheModel}
\noindent By analogy with the characteristics of natural systems, we characterize a \textit{swarm system} as a collection of agents with the following two properties:

\renewenvironment{quote}{\list{}{\leftmargin=0.05\columnwidth\rightmargin=0.05\columnwidth}\item[]}{\endlist}
\begin{quote}
	\textbf{Homogeneity:} All agents in a swarm share a common architecture (\ie\ they have the same dynamics, %
	degrees of freedom and observation capabilities). As such, they are assumed to be interchangeable.\\[0.5\baselineskip] %
	\textbf{Locality:} %
	The agents can observe %
	only %
	parts of the system within a certain range, as determined by their observation capabilities. %
	As a consequence, their decisions depend on their current neighborhood only and not on the whole swarm state. %
\end{quote}

\noindent In principle, any system with these properties can be described %
as a decentralized partially observable Markov decision process (Dec-POMDP) \cite{oliehoek2012decentralized}.
However, the homogeneity property, which turns out to be the key ingredient for scalable inference, is not explicitly captured by this model. Since the number of agents contained in a swarm is typically large, it %
is thus convenient to switch to a 
more compact system representation that exploits the system symmetries.

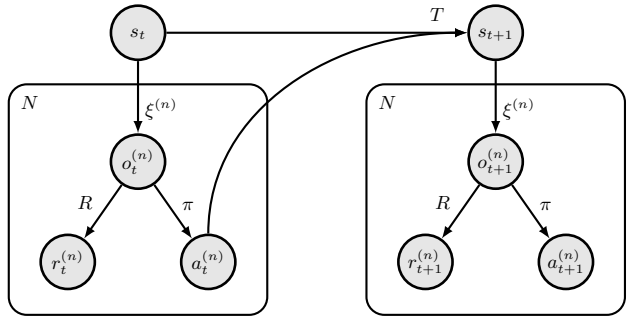
\begin{figure}
	\centering
	\scalebox{0.8}{\begin{tikzpicture}[inner sep=0cm, minimum size = 0.9cm]
		\node (swarm1) [swarm, text depth=0mm] {$s_t$};
		\node (swarm2) [swarm, text depth=0mm, right=5cm of swarm1] {$s_{t+1}$};
		\node (agent1) [agent, below=1.2cm of swarm1] {$o^{(n)}_t$};
		\node (agent2) [agent, below=1.2cm of swarm2] {$o^{(n)}_{t+1}$};
		\node (action1) [action, below right=1cm and 0.5cm of agent1] {$a^{(n)}_t$};
		\node (action2) [action, below right=1cm and 0.5cm of agent2] {$a^{(n)}_{t+1}$};
		\node (reward1) [reward, below left=1cm and 0.5cm of agent1] {$r^{(n)}_t$};
		\node (reward2) [reward, below left=1cm and 0.5cm of agent2] {$r^{(n)}_{t+1}$};
		
		\node (el) [below left=-0.56cm and 1.0cm of swarm2] {};
		
		\node (dummyT1l) [below left=1.7cm and 1.3cm of agent1, minimum size=0cm] {};
		\node (dummyT1r) [below right=1.7cm and 1.3cm of agent1, minimum size=0cm] {};
		\node (dummyT1u) [above=0.3cm of agent1, minimum size=0cm] {};
		\node (dummyT2l) [below left=1.7cm and 1.3cm of agent2, minimum size=0cm] {};
		\node (dummyT2r) [below right=1.7cm and 1.3cm of agent2, minimum size=0cm] {};
		\node (dummyT2u) [above=0.3cm of agent2, minimum size=0cm] {};
		\node (dummyT3lu) [above left=-0.1cm and 1.3cm of swarm1, minimum size=0cm] {};
		\node (dummyT3rd) [below right=-0.1cm and 1.3cm of swarm2, minimum size=0cm] {};
		\node (T1) [template, fit=(dummyT1l) (dummyT1r) (dummyT1u)] {};
		\node (T2) [template, fit=(dummyT2l) (dummyT2r) (dummyT2u)] {};
		\node (N1) [color=black, below right=3mm of T1.north west, minimum size=0cm] {$N$};
		\node (N2) [color=black, below right=3mm of T2.north west, minimum size=0cm] {$N$};

		\draw [-latex, line width=1pt] (swarm1) to node [xshift=2cm, yshift=3mm] {$T$} (swarm2);
		\draw [-latex, line width=1pt] (swarm1) to node [anchor=west, minimum size=0cm, xshift=0.1cm, yshift=-0.2cm] {$\xi^{(n)}$} (agent1);
		\draw [-latex, line width=1pt] (swarm2) to node [anchor=west, minimum size=0cm, xshift=0.1cm, yshift=-0.2cm] {$\xi^{(n)}$} (agent2);
		\draw [-latex, line width=1pt] (agent1) to node [minimum size=0cm, xshift=0.25cm, yshift=0.1cm] {$\pi$} (action1);
		\draw [-latex, line width=1pt] (agent2) to node [minimum size=0cm, xshift=0.25cm, yshift=0.1cm] {$\pi$} (action2);
		\draw [-latex, line width=1pt] (agent1) to node [xshift=-0.3cm, yshift=0.15cm] {$R$} (reward1);
		\draw [-latex, line width=1pt] (agent2) to node [xshift=-0.3cm, yshift=0.15cm] {$R$} (reward2);
		\draw [-latex, line width=1pt, in=180, out = 90] (action1) to (swarm2);
		\end{tikzpicture}}
	\label{subfig:swarMDPmodel}
	\caption{The swarMDP model visualized as a Bayesian network using plate notation. In contrast to a Dec-POMDP, the model explicitly encodes the homogeneity of a swarm. %
	}
	\label{fig:swarMDP}
\end{figure}

For this reason, we introduce a new sub-class of Dec-POMDP models, in the following referred to as \mbox{\textit{swarMDPs}} (\cref{fig:swarMDP}), which explicitly implements a homogeneous agent architecture. %
An agent in this model, which we call a \textit{swarming agent}, is defined as a tuple $\mathbb{A} \defeq ( \mathcal{S}, \mathcal{O} , \mathcal{A}, R, \pi )$, where: 

\begin{itemize}
	
	\item $\mathcal{S},\mathcal{O},\mathcal{A}$ are sets of local states, observations and actions, respectively.
	\item $R:\mathcal{O}\rightarrow\mathbb{R}$ is %
	an agent-level reward function.
	\item $\pi:\mathcal{O}\rightarrow\mathcal{A}$ is the local policy of the agent which later serves as the decentralized control law of %
	the swarm. %
\end{itemize} 

\noindent For the sake of simplicity, we consider only reactive policies in this paper, where $\pi$ is a function of the agent's current observation. Note, however, that the extension to more general policy models (\eg\ belief state policies \cite{kaelbling1998planning} or such that operate on observation histories \cite{oliehoek2012decentralized}) is straightforward. 

With the definition of the swarming agent at hand, we define a swarMDP as a tuple $  ( N, \mathbb{A}, T, \xi ) $, where:
\begin{itemize}
	\item $N$ is the number of agents in the system.
	\item $\mathbb{A} $ is a  swarming agent prototype as defined above.
	\item $ T:\mathcal{S}^N\times \mathcal{A}^N \times \mathcal{S}^N \rightarrow %
	\mathbb{R}$ is the global transition model of the system. Although $T$ is used only implicitly later on, we %
	can access the conditional probability that the system reaches state~$\tilde{s}=(\tilde{s}^{(1)},\ldots,\tilde{s}^{(N)})$ when the agents perform the joint action $a=(a^{(1)},\ldots,a^{(N)})$ at state~$s=(s^{(1)},\ldots,{s}^{(N)})$ as $T(\tilde{s} \given s,a)$, where $s^{(n)}$, $\tilde{s}^{(n)}\in\mathcal{S}$ and $a^{(n)}\in\mathcal{A}$ represent the local states and the local action of agent $n$, respectively.
	\item $\xi:\mathcal{S}^N \rightarrow\mathcal{O}^N$ is the observation model of the system.
\end{itemize}

\noindent The observation model $\xi$ 
tells us which parts of a given system state $s\in\mathcal{S}^N$ %
can be observed by %
whom.  More precisely, $\xi(s) = (\xi^{(1)}(s),\ldots, \xi^{(N)}(s))\in\mathcal{O}^N$ denotes the ordered set of local observations passed on to the agents at %
state~$s$. %
For example, in a school of fish, $\xi^{(n)}$ could be 
the local alignment of a fish to its immediate neighbors (see \cref{section:Vicsek}).
\noindent Note that the agents have no access to their local states $s^{(n)} \in \mathcal{S}$ but only to their local observations $o^{(n)} =\xi^{(n)}(s)\in \mathcal{O}$. %

It should be mentioned that the observation model %
can be also defined locally at the agent level, since the observations are agent-related quantities. However, this would still require a global notion of connectivity between the agents, \eg\ provided in the form of a dynamic graph which defines the time-varying neighborhood of the agents. Using a global observation model, we can %
encode all properties in a single object, yielding a more compact system description. Yet, we need to constrain our model class to those models which respect the homogeneity (and thus the interchangeability) of the agents. %
To be precise, a valid observation model %
needs to ensure that agent~$n$ receives agent $m$'s local observation (and vice versa) if we interchange 
their local states. Mathematically, this means that any permutation of $s \in \mathcal{S}^N$ must result in the same permutation of $\xi(s)$ --~otherwise, the underlying system is not homogeneous. The same property has to hold for the transition model $T$. A generalization to stochastic observations is possible but not considered in~this~paper.

\section{IRL in Swarm Systems}
\label{section:IRL}

\noindent In contrast to existing %
work on IRL, our goal is \textit{not} to develop a new specialized algorithm that %
solves the IRL problem in the swarm case. On the contrary, we show that the homogeneity of our model allows us to reduce the multi-agent IRL problem to %
a single-agent one, for which we can apply a whole class of existing algorithms. This is possible since, %
at its heart, the underlying control problem of the swarMDP is intrinsically a single-agent problem because all agents share the same policy.\footnote{\scriptsize However, the decentralized %
	nature of the problem remains!} In the subsequent sections, we show that this symmetry property also translates to the value functions of the agents.
Algorithmically, %
we exploit the fact that most existing IRL methods, such as \cite{abbeel2004apprenticeship,Neu07,ng2000algorithms,ramachandran2007bayesian,syed2007game,ziebart2008maximum}, %
share a common generic %
form (Algorithm~1),
which involves three main steps \cite{michini2012bayesian}: 1)~policy update 2)~value estimation and 3)~reward update. The important detail to note is that only the first two steps of this procedure are system-specific %
while the third step is, in fact, independent of the target system (see references listed above for details). Consequently, our problem reduces to finding swarm-based solutions for the first two steps such that the overall procedure returns a ``meaningful'' reward model in the IRL context. 
The following sections discuss these steps in detail.
\vspace{-0.7\baselineskip}
\begin{figure}[h]
	\noindent\rule{\columnwidth}{1pt} %
	\textbf{Algorithm 1:} \textsc{Generic IRL} \newline \\[-6.0mm]
	\rule{\columnwidth}{0.2pt}
	\vspace{-1cm}
	\begin{tabbing}
		\hspace{0.35cm} \= \hspace{0.2cm} \= \hspace{2.5cm} \= \kill \\
		\textbf{Input:} expert data $\mathcal{D}$, MDP without reward function \\
		0:	\>Initialize reward function $R^{\lbrace0\rbrace}$ \\
		\> \textbf{for} $i=0,1,2,...$ \\
		1: 	\> 	\> {Policy update:} \> Find optimal policy $\pi^{\lbrace i \rbrace}$ for $R^{\lbrace i \rbrace}$ \\
		2:	\> 	\> {Value estimation:} \> Compute corresponding value $V^{\lbrace i \rbrace}$ \\ %
		3: 	\> 	\> {Reward update:} \> Given $V^{\lbrace i \rbrace}$ and $\mathcal{D}$, compute $R^{\lbrace i+1 \rbrace}$ %
	\end{tabbing}
	\vspace{-\baselineskip}
	\noindent\rule{\columnwidth}{1pt}
\end{figure}

\subsection{Policy Update}
\noindent %
We start with the policy update step, where we are faced with the problem of learning a suitable system policy for a given reward function. For this purpose, we first need to define a suitable learning objective for the swarm setting in the context of the IRL procedure. In the next paragraphs, we show %
that the homogeneity property of our model naturally induces such a learning objective, and we furthermore present a simple learning strategy to optimize this objective.

\subsubsection{Private Value \& Bellman Optimality}
\noindent Analogous to %
the single-agent case \cite{sutton1998reinforcement}, we define the \textit{private value} of an agent $n$ at a swarm state~$s\in\mathcal{S}^N$ under policy~$\pi$ as the expected sum of discounted rewards %
accumulated by the %
agent over time, given that %
all agents execute~$\pi$,
\begin{align}
	V^{(n)}(s \given\pi) &\defeq \Eof{\,\sum\limits_{k=0}^\infty \gamma^k R(\xi^{(n)}(s_{t+k})) \given \pi, s_t = s}{}  \eqkomma
	\label{eq:privateValue}
\end{align}
Herein, $\gamma\in[0,1)$ is a discount factor, and the expectation is with respect to the random system trajectory starting from~$s$. Note that, due to the assumed time-homogeneity of the transition model $T$, the above definition of value is, in fact, independent of any particular starting time~$t$. %
Denoting further by $Q^{(n)}(s,a \given \pi) $ the state-action value of agent~$n$ at state $s$ for the case that 
all agents execute policy $\pi$, except for agent $n$ who performs action $a\in\mathcal{A}$ once and follows $\pi$ thereafter,
we obtain the following Bellman equations:
\begin{align*}
	\label{Eq:Bellman Equations}
	V^{(n)}(s \given \pi) & = R(\xi^{(n)}(s)) + \gamma \sum_{\tilde{s} \in \mathcal{S}^N} \prob(\tilde{s} \given s, \pi ) V^{(n)}(\tilde{s} \given \pi) \eqkomma\\
	Q^{(n)}(s,a \given \pi)&  = R(\xi^{(n)}(s)) + \gamma \sum_{\tilde{s} \in \mathcal{S}^N} \prob^{(n)}(\tilde{s} \given s, a, \pi) V^{(n)}(\tilde{s} \given \pi) \eqpunkt
\end{align*}
Here, $\prob(\tilde{s} \given s, \pi)$ denotes the probability of reaching swarm state $\tilde{s}$ from $s$ when \emph{every} agent performs policy $\pi$ %
and, analogously, $\prob^{(n)}(\tilde{s} \given s, a, \pi)$ denotes the probability of reaching swarm state $\tilde{s}$ from state $s$ if agent $n$ chooses action~$a$ and \emph{all other} agents execute policy $\pi$. 
Note that both these objects are implicitly defined via the transition model $T$.

\subsubsection{Local Value}
\noindent Unfortunately, the value function in \cref{eq:privateValue} is not locally plannable by the agents since they have no access to the global swarm state~$s$.
From a %
control perspective, we thus require an alternative notion of %
optimality that is %
based on local information only and, hence, %
computable by the agents. 
Analogous to the belief value in single-agent systems \cite{ghallab2008exploiting,meuleau1999learning}, we therefore introduce the following \textit{local value function}, %
\begin{equation*}
	V^{(n)}_t(o \given \pi) \defeq \EofP{\prob_t(s\given o^{(n)}=o, \pi)}{V^{(n)}(s \given \pi)} \eqkomma 
\end{equation*}
which represents the expected return of  
agent $n$ under consideration of its current local observation of the global system state. %
In our next proposition, we highlight two key properties of this quantity: %
1) It is not only locally plannable but also reduces the multi-agent problem to a single-agent one in the sense that all local values coincide. 2) In contrast to the private value, %
the local value is time-dependent because the conditional probabilities $\prob_t(s\given o^{(n)}=o, \pi)$, in general, depend on time. However, it converges to a stationary value asymptotically under suitable conditions.

\begin{MyProposition}
	\label{Proposition:Limiting Value}
	Consider a swarMDP as defined above and %
	the stochastic process $\lbrace S_t\rbrace_{t=0}^\infty$ of the swarm state induced by %
	the %
	system policy $\pi$. If the initial state distribution %
	of the system is invariant under permutation\footnote{\scriptsize Since we assume that the agents are interchangeable, it %
		follows naturally to consider only %
		permutation-invariant initial distributions.} of agents, %
	all local value functions %
	are identical, %
	\begin{equation}
	V^{(m)}_t(o\given\pi) = V^{(n)}_t(o\given\pi) \quad \forall m,n \eqpunkt
	\label{eq:LocalValue}
	\end{equation}
	In this case, we may drop the agent index and denote the common local value function as $V_t(o \given \pi)$. If, furthermore, it holds that $ S_t \xrightarrow{a.s.} S$ %
	for some $S$ with law  $P$ and the common local value function is continuous almost everywhere (\ie\ its set of discontinuity points is $P$-null) and bounded above, then  %
	the local value function $V_t(o \given \pi)$ will converge to a limit,
	\begin{equation}
	V_t(o \given \pi) \rightarrow  V(o \given \pi) \eqkomma
	\label{eq:stationaryValue}
	\end{equation}
	where $V(o \given \pi)= \EofP{\prob(s \given o^{(n)}=o,\pi)}{V^{(n)}(s \given \pi) } $. %
\end{MyProposition}

\begin{figure*}
	\centering
	\vspace{0.2cm}
	\setlength{\fboxsep}{0cm}
	\begin{subfigure}{0.22\textwidth}
		\centering
		\fbox{\includegraphics[width = 3.5cm]{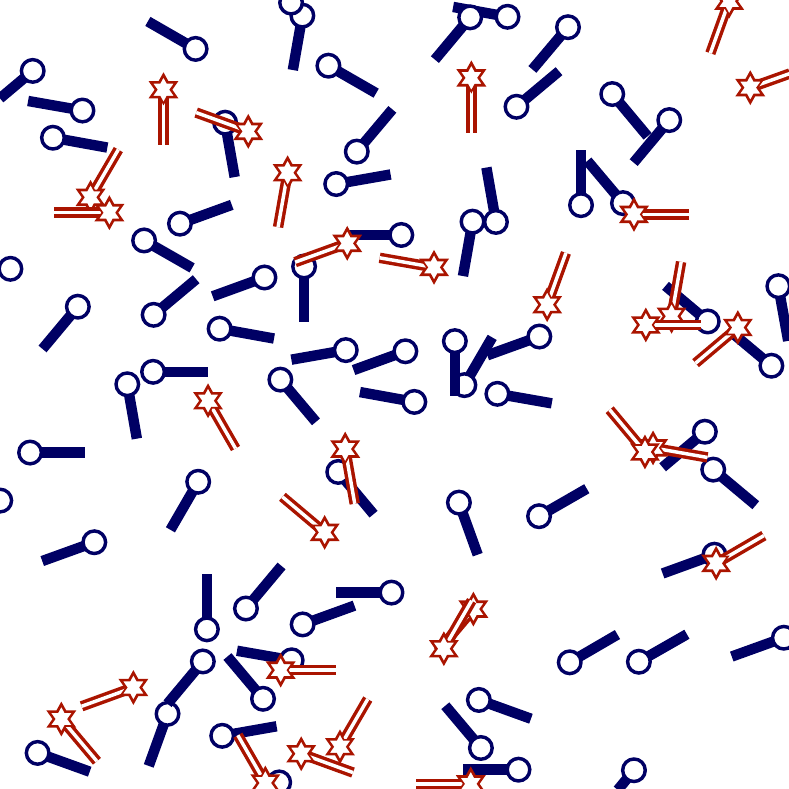}}
	\end{subfigure}
	\hspace{0.2cm}
	\begin{subfigure}{0.22\textwidth}
		\centering
		\fbox{\includegraphics[width = 3.5cm]{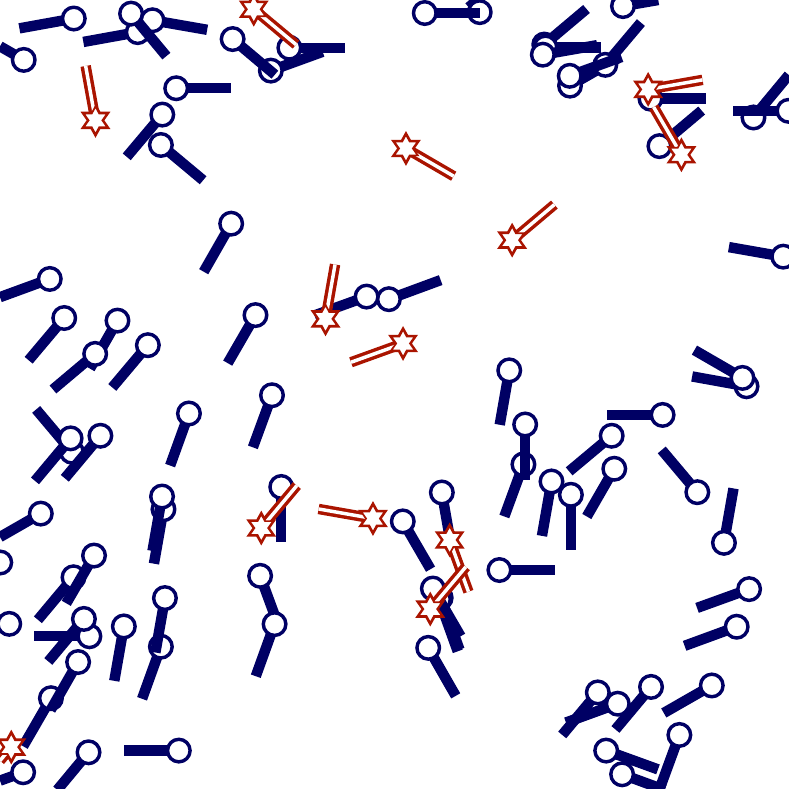}}
	\end{subfigure}
	\hspace{0.2cm}
	\begin{subfigure}{0.22\textwidth}
		\centering
		\fbox{\includegraphics[width = 3.5cm]{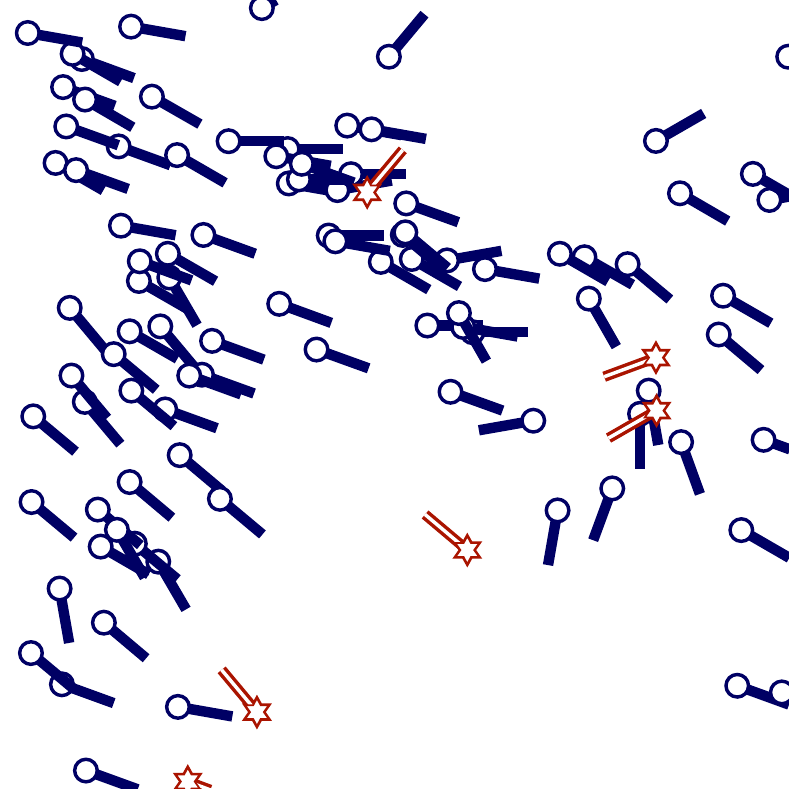}}
	\end{subfigure}
	\hspace{0.2cm}
	\begin{subfigure}{0.22\textwidth}
		\centering
		\fbox{\includegraphics[width = 3.5cm]{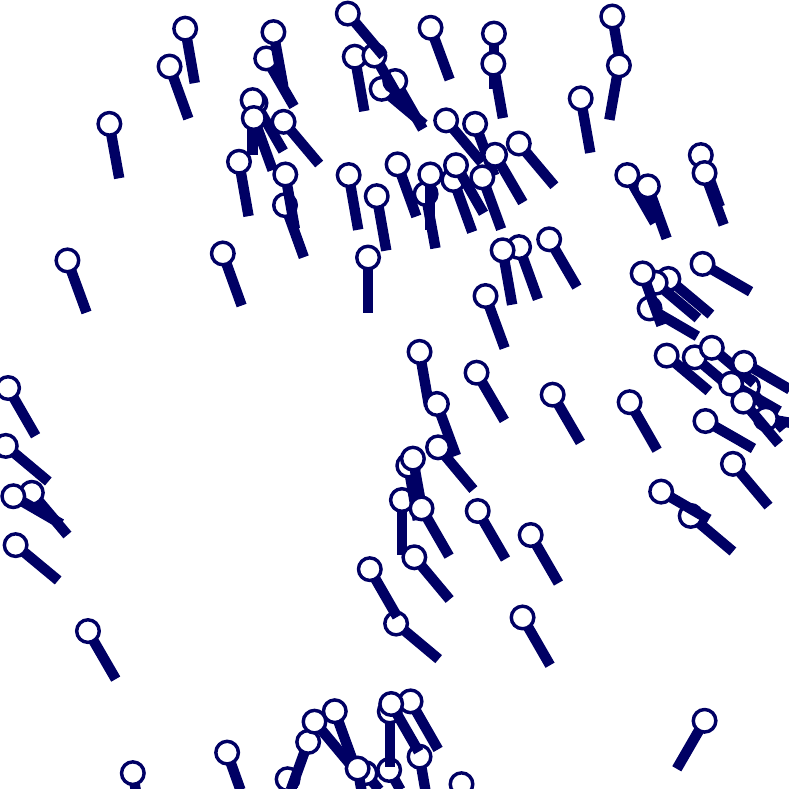}}
	\end{subfigure}
	
	\vspace{0.2cm}
	
	\begin{tikzpicture}[auto]
	\draw [-latex, line width=1pt] (0,0) to node [swap] {learning iterations} (15,0);
	\end{tikzpicture}
	\caption{Snapshots of the proposed learning scheme applied to the Vicsek model (\cref{section:Vicsek}). The agents are divided into a greedy set~(\hspace{0.1mm}{\protect \tikz \protect \node [minimum size=0cm, inner sep=-0.02cm] {\protect \includegraphics{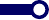}};}) and an exploration set (\hspace{0.3mm}{\protect \tikz \protect \node [minimum size=0cm, inner sep=-0.045cm] {\protect \includegraphics{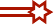}};}\hspace{0.1mm}) %
		to facilitate the exploration %
		of locally desynchronized swarm states. The size of the exploration set is reduced over time to gradually transfer the system into a homogeneous stationary behavior.}%
	\label{fig:Cooling}
\end{figure*}

\begin{proof}[Proof]
	
	Fix any two agents, say agent 1 and 2. For these agents, define a permutation operation $\sigma: \mathcal{S}^N \rightarrow \mathcal{S}^N $ as
	\begin{equation*}
		\sigma (s) \defeq (s^{(2)},s^{(1)},s^{(3)},\ldots, s^{(N)}) \eqkomma
	\end{equation*}
	where $s=(s^{(1)},s^{(2)},s^{(3)},\ldots, s^{(N)})$. %
	Due to the homogeneity of the system, \ie\ since $R(\xi^{(1)}(s)) = R(\xi^{(2)}(\sigma(s)))$ and \mbox{$\prob(\tilde{s} \given s, \pi) = \prob(\sigma(\tilde{s}) \given \sigma(s), \pi)$}, it follows immediately that $V^{(1)}(s\given\pi) = V^{(2)}(\sigma(s)\given\pi) \ \forall s$.
	This essentially means: the value assigned to agent 1 at swarm state $s$ is the same as the value that would be assigned to agent 2 %
	if we interchanged their local states, \ie\ at state $\sigma(s)$. %
	Note that this is effectively the same as renaming the agents. The homogeneity of the system ensures that the symmetry of the initial state distribution~$\prob_0(s)$ is maintained at all subsequent points in time, \ie\ $\prob_t(s \given \pi)=\prob_t(\sigma(s) \given \pi)\ \forall s,t$. In particular, it holds that $\prob_t(s\given o^{(1)}=o, \pi)=\prob_t(\sigma(s) \given o^{(2)}=o, \pi)\ \forall s,t$ and, accordingly, %
	\begin{align*}
		&	V_t^{(1)}(o \given \pi) - V_t^{(2)}(o \given \pi) \\
		&= \EofP{\prob_t(s \given o^{(1)}=o, \pi)}{V^{(1)}(s \given \pi) } -\EofP{\prob_t(s \given o^{(2)}=o, \pi)}{V^{(2)}(s \given \pi) } \\
		&= \sum\limits_{s\in\mathcal{S}^N} \Big(\prob_t(s \given o^{(1)}=o, \pi) V^{(1)}(s \given \pi) \ \ldots \\[-2mm]
		&\phantom{=}\hspace{10mm} \ldots \ -\ \prob_t(\sigma(s) \given o^{(2)}=o, \pi) V^{(2)}(\sigma(s) \given \pi) \Big)=0 \eqkomma
	\end{align*}
	which shows that the local value functions are identical for all agents. 
	Treating the value as a random variable and using the fact that it is continuous almost everywhere, it follows that $V^{(n)}(S_t \given \pi) \xrightarrow{a.s.} V^{(n)}(S \given \pi)$ since  $S_t \xrightarrow{a.s.} S$. As we %
	assume the function to be finite, \ie\ \mbox{$| V^{(n)}(S_t \given \pi) | < V^*$} for some $V^* \in \mathbb{R}$, it holds by conditional
	dominated convergence theorem~\cite{billingsley2013convergence} that $\Eof{V^{(n)}(S_t\given\pi) \given o^{(n)}=o, \pi} \rightarrow \Eof{V^{(n)}(S\given\pi) \given o^{(n)}=o, \pi} $, \ie\ $ V_t(o \given \pi) \rightarrow  V(o \given \pi)$. 
\end{proof}
\subsubsection{Heterogeneous Q-learning}
\label{section:forwardProblem}

\noindent With the local value in \cref{eq:LocalValue}, we have introduced a system-wide performance measure which can be evaluated at the agent level and, hence, can be used by the agents for local planning. Yet, its computation involves the evaluation of an expectation with respect to the current swarm state of the system. This %
requires the agents to maintain a belief about the global system state %
at any point in time to %
coordinate their actions, which itself is a hard problem.\footnote{\scriptsize In principle, this is possible since -- in contrast to a Dec-POMDP -- each agent knows the policy of the other agents.} %
However, for many swarm-related tasks (\eg\ consensus problems \cite{ren2005survey}), %
it is sufficient to optimize 
the stationary behavior of the system. This is, in fact, %
a much easier task since it allows us to forget about the temporal aspect of the problem. 

In this section, we present a comparably simple learning method, specifically tailored to the swarm setting, which solves this task by optimizing the system's stationary value in \cref{eq:stationaryValue}.  %
Similar to the local value function, we start by defining a \textit{local Q-function} for each agent, %
\begin{equation*}
	Q_t^{(n)}(o,a\given \pi) \defeq \EofP{\prob_t(s\given o^{(n)}=o, \pi)}{Q^{(n)}(s,a \given \pi)} \eqkomma \\
\end{equation*}
which assesses the quality of a particular action played by agent $n$ at time $t$. %
Following the same line of argument as %
before, one can show that these Q-functions are again identical for all agents and, moreover, that they converge to the following asymptotic value function,
\begin{equation}
Q(o,a \given \pi) =\EofP{\prob(s \given o^{(n)}=o, \pi)}{Q^{(n)}(s,a \given \pi) } \eqkomma
\label{eq:localQ}
\end{equation}
which can be understood as the state-action value of a generic agent that is coupled to a stationary field generated by and executing policy $\pi$. %
In the following, we pose the task of optimizing this Q-function as a game-theoretic one. To be precise, we consider a hypothetical game between each agent and the environment surrounding it, where the agent plays the optimal response to this stationary field, %
\begin{equation*}
	\pi_\text{R}(o \given \pi) \defeq \argmax\limits_{a\in\mathcal{A}}Q(o,a \given \pi) \eqkomma
\end{equation*}
and the environment reacts with a new swarm behavior generated by this policy.
By definition, any optimal system policy $\pi^*$ describes a fixed-point of this game,
\begin{equation*}
	\pi_\text{R}(o \given \pi^*) = \pi^*(o) \eqkomma
\end{equation*}
which motivates the following iterative learning scheme: \linebreak Starting with an arbitrary initial policy, we run the system until it reaches its stationary behavior and estimate the corresponding asymptotic Q-function. Based on this Q-function, we update the system policy according to the best response operator defined above. The updated policy, in turn, induces a new swarm behavior for which we estimate a new Q-function, and so on. %
As soon as we reach a fixed-point, the system has arrived at %
an optimal behavior in the form of a symmetric Nash equilibrium where all agents collectively execute a policy which, for each agent individually, provides the optimal response to the other agents' behavior. 

However, the following practical problems remain: 1) In general, it can be time-consuming to wait for the system to reach its stationary behavior at each iteration of the algorithm. 2) At stationarity, we need a way to estimate the corresponding stationary 
Q-function. Note that this involves both estimating the Q-values of %
actions that are dictated by the current policy as well as Q-values of actions that deviate from %
the current behavior, %
which requires a certain amount of exploratory %
moves. 
As a solution to both problems, we propose the following \textit{heterogeneous learning scheme}, which artificially breaks the symmetry of the system %
by separating the 
agents into two disjoint groups: a greedy set and an exploration set. While the agents in the greedy set provide a reference behavior in the form of the optimal response to the current Q-function %
shared between all agents, the agents in the exploration set randomly explore the quality of different actions in the context of the current system policy. %
At each iteration, the gathered experience of all agents is processed sequentially via the following Q-update \cite{watkins1992q}, 
\begin{equation*}
	\hat{Q}(o_t^{(n)},a_t^{(n)}) \leftarrow (1-\alpha) \hat{Q}(o_t^{(n)},a_t^{(n)}) + \alpha \big( r_t^{(n)} + \gamma \max_{a\in\mathcal{A}} \hat{Q}(o_{t+1}^{(n)}, a) \big) \eqkomma
\end{equation*}
with learning rate $\alpha\in(0,1)$.
Over time, more and more exploring agents are assigned to the greedy set %
so that the system is gradually transferred into a \textit{homogeneous stationary} regime and thereby smoothly guided towards a fixed-point policy (see \Cref{fig:Cooling}).
Herein, the learning rate $\alpha$ naturally %
reduces the %
influence of experience acquired at early (non-synchronized) stages of the %
system, which allows us to update the system policy %
without having to wait until the swarm converges to its stationary behavior.

The heterogeneity %
of the system during the learning phase ensures that also locally desynchronized swarm states %
are well-explored (together with their local Q-values) %
so that the agents can learn adequate responses to out-of-equilibrium situations. This phenomenon is best illustrated by the agent constellation in third sub-figure of \Cref{fig:Cooling}. It shows a situation that is highly unlikely %
under a homogeneous learning scheme as it requires a series of consecutive exploration steps by only a few agents %
while all their neighbors need to behave consistently optimally at the same time.
The final procedure, which can be interpreted as a model-free variant of policy iteration \cite{lagoudakis2003least} in a non-stationary environment, is summarized in Algorithm~2, together with a pictorial description of the main steps in \cref{fig:learningScheme}. %
While we cannot provide a convergence proof at this stage, the algorithm converged in all our simulations and generated policies with a performance close to that of the expert system (see \cref{section:results}).

\begin{figure}
	\centering
	\vspace{0.1cm}
	\scalebox{0.8}{
		\begin{tikzpicture}[minimum size = 1cm, line width = 1pt,auto]
		\node [block] (Q) at (0,0) {$Q(o,a \given \pi)$};
		\node [block, right=1cm of Q] (pi) {$\pi(o)$};
		\node [block, right=1cm of pi] (samples) {$\left\lbrace\lbrace o^{(n)}_t, a^{(n)}_t, r^{(n)}_t, o^{(n)}_{t+1} \rbrace_{n=1}^N\right\rbrace_t$};
		\draw [-latex] (Q) to node [yshift=-0.2cm] {(a)} (pi);
		\draw [-latex] (pi) to node [yshift=-0.2cm] {(b)} (samples);
		
		\draw (samples.east) [rounded corners] -- ++(0.6,0) [rounded corners] -- ++(0,1.2) [rounded corners] -- node [yshift=0.2cm] {(c)} ++(-10.5,0) [rounded corners] -- ++(0,-1.2) [rounded corners, -latex] -- (Q.west) ;
		
		\end{tikzpicture}}
	\caption{Pictorial description of the proposed learning scheme. (a)~The next policy is obtained from the current estimate of the system's stationary Q-function. (b)~Heterogeneous one-step transition %
		of the system. (c)~The estimate of the Q-function is updated based on the new experience.} %
	\label{fig:learningScheme}
\end{figure}
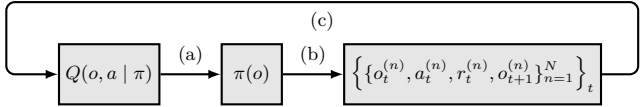

\begin{figure}[]
	\noindent\rule{\columnwidth}{1pt} %
	\textbf{Algorithm 2:} \textsc{Heterogeneous Q-Learning} \newline \\[-6.0mm]
	\rule{\columnwidth}{0.2pt}
	\vspace{-1cm}
	\begin{tabbing}
		\hspace{0.35cm} \= \hspace{0.25cm} \= \kill \\
		\textbf{Input:} swarMDP without policy $\pi$ \\
		0:	\> Initialize shared Q-function, learning rate and fraction \\
		\> of exploring agents (called the temperature) \\
		\> \textbf{for} $i=0,1,2,...$ \\
		1:	\>  \> Separate the swarm into exploring and greedy agents \\
		\>  \> according to the current temperature\\
		2:	\>  \> Based on the current swarm state and Q-function, \\
		\>  \> select actions for all agents \\
		3:	\>  \> Iterate the system and collect rewards \\
		4:	\>  \> Update the Q-function based on the new experience \\
		5:	\>  \> Decrease the learning rate and the temperature 
	\end{tabbing}
	\vspace{-\baselineskip}
	\noindent\rule{\columnwidth}{1pt}
\end{figure}

\subsection{Value Estimation}
\noindent In the last section, we have shown a way to implement the policy update in Algorithm~1 based on local %
information %
acquired at the agent level. %
Next, we need to assign a suitable value to the obtained policy %
which allows a comparison to the expert behavior in the subsequent reward update step.

\begin{figure*}
	\centering
	\hspace{-0.0cm}
	\begin{subfigure}{12.5cm}
		\begin{tikzpicture}
		\node at (0,0) {\includegraphics[width=5.4cm]{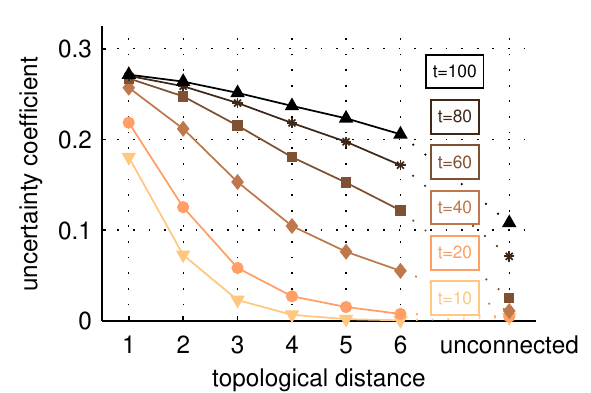}};
		\node at (6.3,0) {\includegraphics[width=7.2cm]{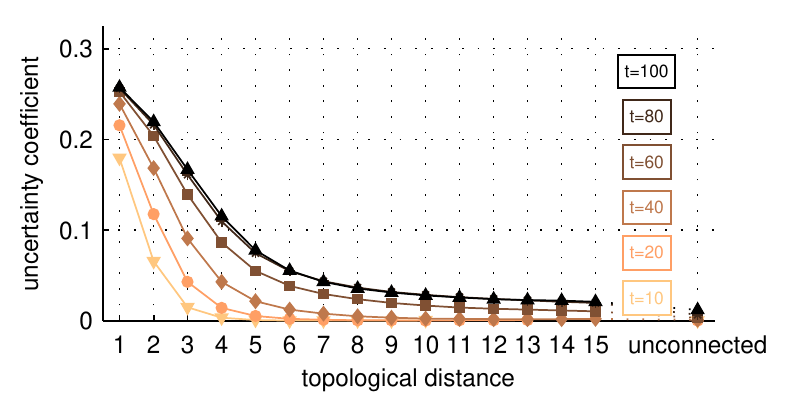}};
		\end{tikzpicture}
		\caption{Uncertainty coefficient for the interaction radii $\rho=0.125$ (left) and $\rho=0.05$ (right). The values are based on a kernel density estimate of the joint distribution of two agents' headings.}
		\label{fig:redundancy}
	\end{subfigure}
	\hspace{0.6cm}	
	\begin{subfigure}{4cm}
		\scalebox{0.7}{\begin{tikzpicture}
			\node at (0:-0.1) {\includegraphics[width=5cm]{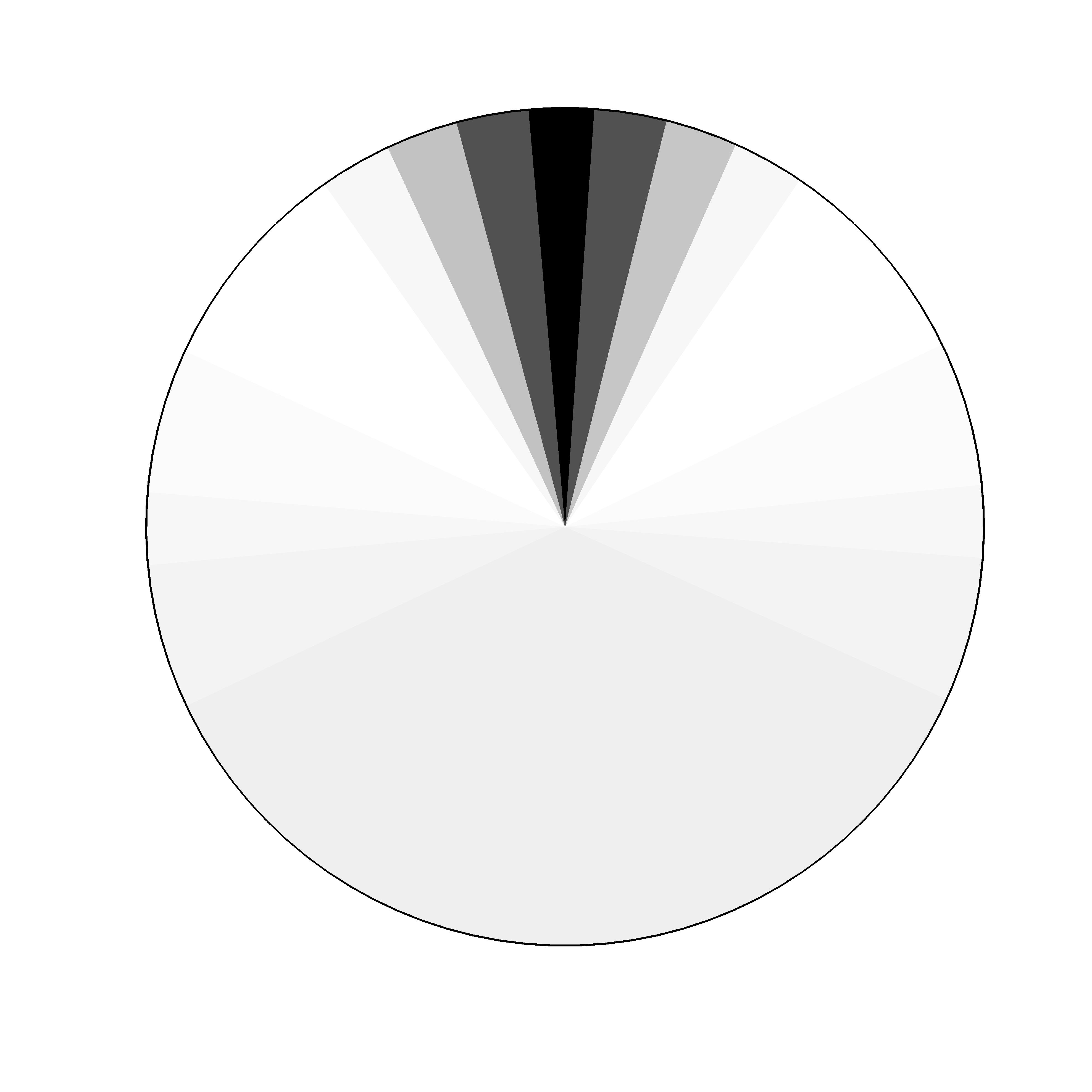}};
			
			\foreach \x in {-150,-120,...,-30}
			\node at (360-\x+90:2.5) {$\x^\circ$};
			\foreach \x in {150,120,...,30}
			\node at (360-\x+90:2.5) {$+\x^\circ$};
			
			\node at (270:2.5) {$\pm180^\circ$};
			\node at (90:2.5) {$\pm0^\circ$};
			
			\end{tikzpicture}}
		\caption{Learned reward model.}
		\label{fig:rewardVicsek}
	\end{subfigure}
	\caption{Simulation results for the Vicsek model. (a) The system's uncertainty coefficient as a function of the topological distance between two agents, estimated from 10000 Monte Carlo runs of the expert system. (b) Learned reward model as a function of an agent's local misalignment, averaged over 100 Monte Carlo experiments. Dark color indicates high reward.}
	\label{fig:VicsekResults}
\end{figure*}
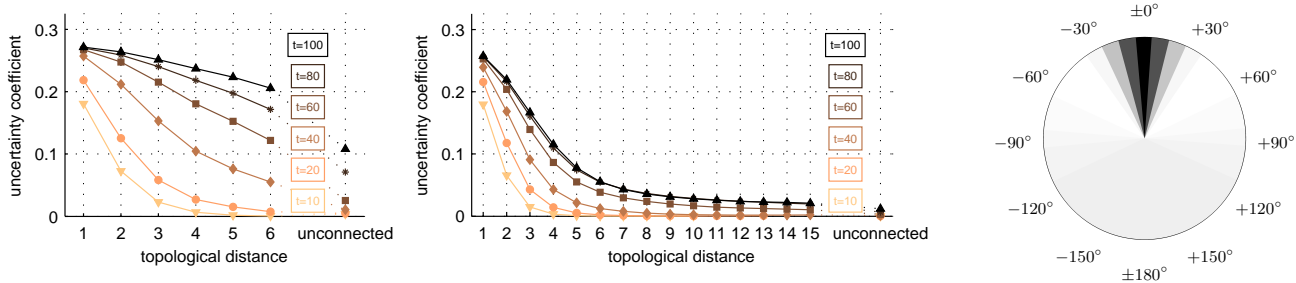

\subsubsection{Global Value}
\label{section:valueEstimation}
\noindent The comparison of the learned behavior and the expert behavior should take place on a global level, since we want the updated reward function to cause a new system behavior which mimics the expert behavior \textit{globally}. %
Therefore, %
we introduce the following \textit{global value},
\begin{equation*}
	V^{(n)}_{|\pi} \defeq \EofP{\prob_0(s)}{V^{(n)}(s \given \pi)} \eqkomma
\end{equation*}
which represents the expected return of an agent under $\pi$, averaged over all possible initial %
states of the swarm.
From the system symmetry, \ie\ since $\prob_0(s)=\prob_0(\sigma(s))$, it follows immediately that this global value is independent of the specific agent under consideration,
\begin{gather*}
	V^{(m)}_{|\pi} = V^{(n)}_{|\pi} \quad \forall m,n \eqpunkt
\end{gather*}
Hence, %
the global value should be considered as a system-related performance measure (as opposed to an agent-specific property), which may be utilized for the reward update in the last step of the algorithm. We can construct an unbiased estimator for this quantity from any local agent trajectory, %
\begin{equation}
\label{eq:ValueEstimateSingle}
\hat{V}_{|\pi}^{(n)} = \sum\limits_{t=0}^{\infty}\gamma^tR(\xi^{(n)}(s_t)) = \sum\limits_{t=0}^{\infty}\gamma^tr_t^{(n)}\eqpunkt
\end{equation}
Since all local estimators are identically distributed, %
we can increase the accuracy of our estimate by considering the information provided by the whole swarm,
\begin{equation}
\label{eq:ValueEstimate}
\hat{V}_{|\pi} = \frac{1}{N}\sum\limits_{n=1}^{N} \hat{V}_{|\pi}^{(n)} = \frac{1}{N}\sum\limits_{n=1}^{N}\sum\limits_{t=0}^{\infty}\gamma^tr_t^{(n)}\eqpunkt
\end{equation}
Note, however, %
that the local estimators are not independent since all agents are correlated through the system process. Nevertheless, due to the local coupling structure of a swarm, this correlation is caused only \textit{locally}, which means that the correlation between any two agents will drop when their topological distance increases. We demonstrate this phenomenon for the Vicsek model in \cref{section:Vicsek}.

\subsection{Reward Update}
\label{sec:rewardUpdate}
\noindent The last step %
of Algorithm~1 consists in updating the estimated reward function. %
Depending on the single-agent IRL framework in use, this involves an algorithm-specific optimization procedure, \eg\ in the form of a quadratic program \cite{abbeel2004apprenticeship,ng2000algorithms} or a gradient-based optimization \cite{Neu07,ziebart2008maximum}. For our experiments in \cref{section:results}, we follow the max-margin approach presented in \cite{abbeel2004apprenticeship}; %
however, the procedure can be replaced with other value-based methods (see \cref{section:IRL}). %

For this purpose, the local %
reward function is represented as a linear combination of observational features, $R(o)=w^\top\phi(o)$, with weights $w\in\mathbb{R}^d$ and a given feature function $\phi:\mathcal{O}\rightarrow\mathbb{R}^d$.
The feature weights after the $i$th iteration of Algorithm~1 
are then obtained as
\begin{equation*}
	w^{\{i+1\}} = \argmax_{w:||w||_2 \leq 1} \ \underset{j\in\{1,\ldots,i\}}{\text{min }} \ w^\top (\mu_E - \mu^{(j)}) \eqpunkt
\end{equation*}
\nopagebreak
where $\mu_E$ and $\{\mu^{(j)}\}_{j=1}^i$ are the \textit{feature expectations} \cite{abbeel2004apprenticeship} of the expert policy and the learned policies up to iteration~$i$. Simulating a one-shot learning experiment, we estimate these quantities from a single system trajectory based on \cref{eq:ValueEstimate},
\begin{equation*}
	\hat{\mu}(\pi) = \frac{1}{N}\sum\limits_{n=1}^{N}\sum\limits_{t=0}^{\infty}\gamma^t \phi(\xi^{(n)}(s_t))\eqpunkt
\end{equation*}
where the state sequence $(s_0, s_1, s_2, \ldots )$ is generated using the respective policy $\pi$. For more details, we refer to \cite{abbeel2004apprenticeship}.

\section{Simulation Results}
\label{section:results}
\noindent In this section, we provide simulation results for two different system types. 
For the policy update, the initial %
number of exploring agents is set to 50\% of the population size and the learning rate is initialized close to 1. Both quantities are controlled by a quadratic decay which ensures that, at the end of the learning period, \ie\ after 200 iterations, %
the learning rate reaches zero and there are no exploring agents left.
Note that these parameters are by no means optimized; yet, in our experiments we observed that the learning results are largely insensitive to 
the particular choice of values. %
Since the agents' observation space is one-dimensional in both experiments, we use a simple tabular representation for the learned Q-function; for higher-dimensional problems, one needs to resort to function approximation \cite{lagoudakis2003least}. Videos can be found at \url{http://www.spg.tu-darmstadt.de/aamas2017}.

\subsection{The Vicsek Model}
\label{section:Vicsek}
\noindent %
First, we test our framework on the Vicsek model of self-propelled particles \cite{vicsek1995novel}. The model consists of a fixed number of particles, or agents, living in the unit square $[0,1]\times[0,1]$ with periodic boundary conditions. Each agent $n$ moves with a constant absolute velocity $v$ and is characterized by its location $x^{(n)}_t$ and orientation $\theta^{(n)}_t$ in the plane, as summarized by the local state variable $s^{(n)}_t\defeq(x^{(n)}_t,\theta^{(n)}_t)$.
The time-varying neighborhood structure of the agents is determined by a fixed interaction radius $\rho$.
At each time instance, the agents' orientations get synchronously updated to the average orientation of their neighbors (including themselves) with additive random perturbations $\{\Delta\theta^{(n)}_t\}$,
\begin{equation}
\begin{aligned}
\theta^{(n)}_{t+1} &= \langle\theta^{(n)}_t\rangle_\rho + \Delta\theta^{(n)}_t \eqkomma\\
x^{(n)}_{t+1} &= x^{(n)}_t + v^{(n)}_t \eqpunkt
\end{aligned}
\label{eq:VicsekOriginal}
\end{equation}
Herein, $\langle\theta^{(n)}_t\rangle_\rho$ denotes the %
mean orientation of all agents within the $\rho$-neighborhood of agent $n$ at time $t$, and $v^{(n)}_t=v\cdot[\cos \theta^{(n)}_t,\ \sin \theta^{(n)}_t]$ is the velocity vector of agent $n$.

\begin{figure*}
	\centering
	\begin{minipage}[b]{0.997\columnwidth}
		\centering
		\setlength{\fboxsep}{0cm}
		\begin{tikzpicture}
		\node (a) at (0,0) {\begin{subfigure}{\sizeA}
			\centering
			\fbox{\includegraphics[width=3.0cm]{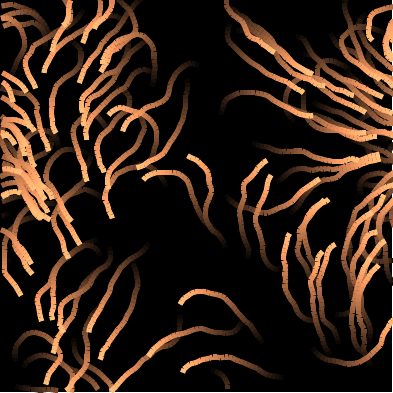}}
			\end{subfigure}};
		\node (c) [right = -2.0cm of a] {\begin{subfigure}{\sizeB}
			\centering
			\fbox{\includegraphics[width=3.0cm]{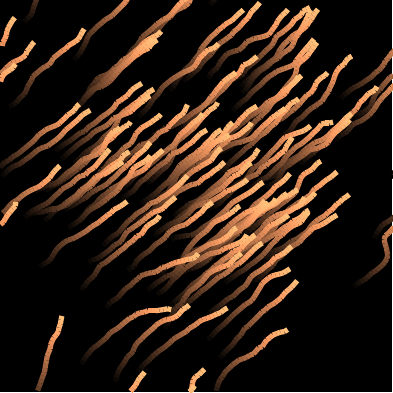}}
			\end{subfigure}};
		\node (d) [below = 0.0cm of a] {\begin{subfigure}{\sizeA}
			\centering
			\fbox{\includegraphics[width=3.0cm]{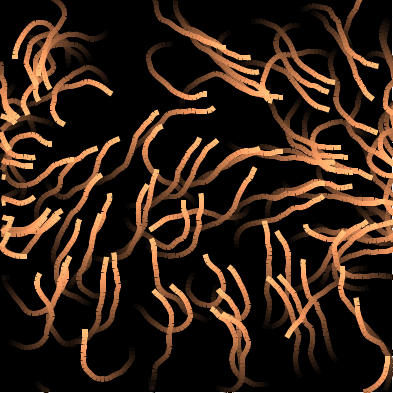}}
			\end{subfigure}};
		\node (f) [right = -2.0cm of d] {\begin{subfigure}{\sizeB}
			\centering
			\fbox{\includegraphics[width=3.0cm]{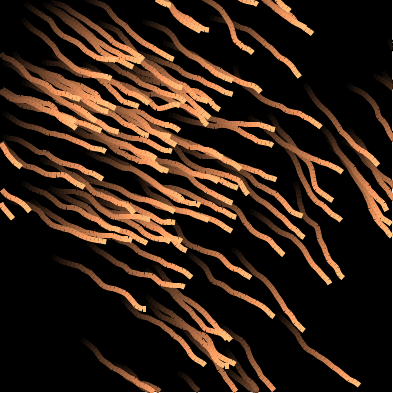}}
			\end{subfigure}};

		\node (trans) [above left=-1.65cm and 0.0cm of a, anchor=west] {\small optimal};
		\node (stat) [above left=-1.65cm and 0.0cm of d, anchor=west]  {\small learned \phantom{p}};
		\node (optimal) [above=0cm of a, text depth=0ex] {transient};
		\node (learned) [above=0cm of c, text depth=0ex] {stationary};
		\node (r1) [above right=-1.35cm and -0.9cm of c, anchor=north]  {};
		\node (r2) [above right=-1.35cm and -0.9cm of f, anchor=north]  {};
		
		\begin{pgfonlayer}{background}
		\node [connect, fit=(trans) (r1)] {};
		\node [connect, fit=(stat) (r2)] {};
		\end{pgfonlayer}
		
		\end{tikzpicture}
		\caption{Illustrative trajectories of the Vicsek model generated under the optimal policy and a learned policy. A color-coding scheme is used to indicate the temporal progress. %
		}
		\label{fig:policies}
	\end{minipage}
	\hspace{0.75\columnsep}
	\begin{minipage}[b]{0.997\columnwidth}
		\centering
		\includegraphics[scale=0.8]{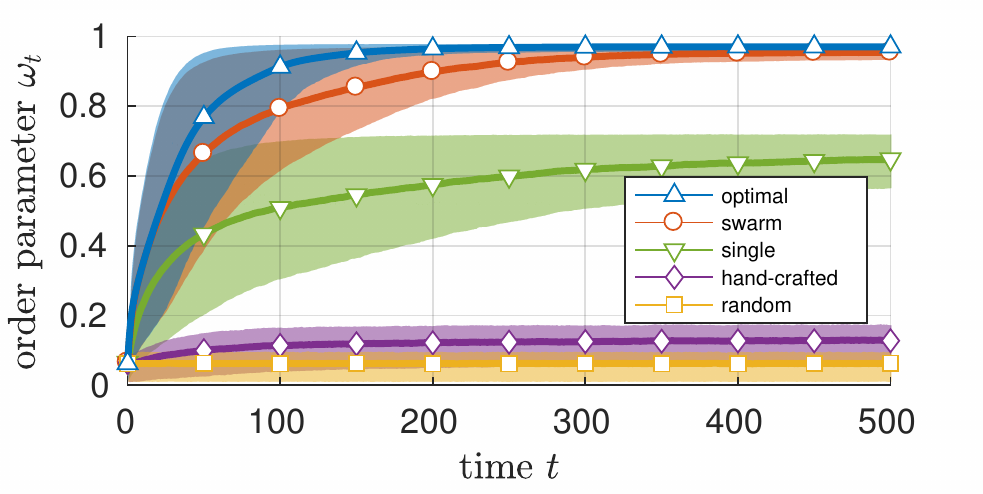}
		\caption{Slopes of the order parameter $\omega_t$ in the Vicsek model. %
			From top to bottom, the curves show the results for the expert policy, the learned IRL policy, %
			the result we get when the %
			feature expectations are estimated from just one single agent, for a hand-crafted reward function, and for random policies. For the optimal policy, we show the empirical mean and the corresponding 10\% and 90\% quantiles, based on 10000 Monte Carlo runs. For the learned policies, we instead show the average over 100 conditional quantiles (since the outcome of the learning process is random), each based on 100 Monte Carlo runs with a fixed policy.}
		\label{fig:orderParameter}
	\end{minipage}
\end{figure*}

Our goal is to learn a model %
for this expert behavior from recorded agent trajectories using the proposed framework. %
As a simple observation mechanism, we let the agents in our model compute the angular distance to the average orientation of their neighbors, \ie\ \mbox{$o^{(n)}_t=\xi^{(n)}(s_t)\defeq\langle\theta^{(n)}_t\rangle_\rho-\theta^{(n)}_t$}, giving them the ability to monitor their local misalignment. 
For simplicity, %
we discretize the observation space $[0,2\pi)$ into 36 equally-sized intervals (\cref{fig:orderParameter}), corresponding to the features $\phi$ (\cref{sec:rewardUpdate}). 
Furthermore, we coarse-grain the space of possible direction changes to $[-60^\circ, -50^\circ, \dots, 60^\circ]$, resulting in a total of $13$~actions available to the agents. %
For the experiment, we use a system size of $N=200$, an interaction radius of $\rho=0.1$ (if not stated otherwise), an absolute velocity of $v=0.1$, a discount factor of $\gamma=0.9$, and a zero-mean Gaussian noise model for %
$\{\Delta\theta^{(n)}_t\}$ with a standard deviation of $10^\circ$. These parameter values are chosen such that the expert system operates in an ordered phase~\cite{vicsek1995novel}.

\subsubsection*{Local Coupling \& Redundancy}
In \Cref{section:valueEstimation} we claimed that, due to the local coupling in a swarm, 
the correlation between any two agents will decrease with growing topological distance.
In this section, we substantiate our claim by analyzing %
the coupling strength in the system %
as a function of the topological distance between the agents. %
As a measure of (in-)dependence, we employ the \textit{uncertainty coefficient} \cite{press2007numerical}, a normalized version of the mutual information, %
which %
reflects the %
amount of information we can predict about an agent's orientation by observing that of another agent. As opposed to %
linear correlation, this measure is 
able to capture non-linear dependencies and is, hence, more meaningful in the context of the Vicsek model whose %
state dynamics are inherently non-linear.

\Cref{fig:VicsekResults}(a) depicts the result of our analysis which nicely reveals the spatio-temporal flow of information in the system. %
It confirms that the mutual information exchange between the agents strongly depends on the strength of their coupling which is determined by 1) their topological distance and 2) the number of connecting links %
(seen from the fact that, for a fixed distance, the dependence grows with the interaction radius). We also see that, for increasing radii, the dependence grows even for agents that are temporarily not connected through the system, %
due to the increasing chances of having been connected at some earlier stage.

\subsubsection*{Learning Results}
An inherent problem with any IRL approach is the assessment of the extracted reward function as there is typically no ground truth to compare with. 
The simplest way to check the plausibility of the result is by subjective inspection: since a system's reward function  can be regarded as a concise description of the task being performed, %
the estimate should explain the observed system behavior reasonably well.
As we can see from \Cref{fig:VicsekResults}(b), this is indeed the case for the obtained result. Although %
there is no ``true'' reward model for the Vicsek system, we can %
see from the system equations in \eqref{eq:VicsekOriginal} that %
the agents tend to align over time. %
Clearly, such dynamics can be induced by giving higher rewards for %
synchronized states and lower (or negative) rewards for misalignment.
Inspecting further the induced system dynamics (\cref{fig:policies}), %
we observe that the algorithm is %
able to %
reproduce the behavior of the expert system, both during the transient phase and at stationarity. Note that the absolute direction of travel is not important here as the model considers only relative angles between the agents.
Finally, we compare the results in terms of the %
\textit{order parameter} \cite{vicsek1995novel}, which provides a measure for the total alignment of the swarm,
\begin{equation*}
	\omega_t \defeq \frac{1}{Nv}\left\vert\sum\limits_{n=1}^Nv^{(n)}_t\right\vert \in [0,1]\eqkomma
\end{equation*}
with values close to 1 indicating strong synchronization of the system.
\Cref{fig:orderParameter} depicts its slope for different system policies, including the expert policy and the learned ones. %
From the result, we can see a considerable performance gain for the proposed value estimation scheme (\cref{eq:ValueEstimate}) as compared to a single-agent approach (\cref{eq:ValueEstimateSingle}). This again confirms our findings from the previous section since the increase in performance %
has to stem from the additional information provided by the other agents. As a further reference, %
we also show the result for a hand-crafted reward model, where we provide a positive reward only if the local observation of an agent falls in the discretization interval centered around $0^\circ$ misalignment. %
As we can see, the learned reward model %
significantly outperforms the ad-hoc solution.

\begin{figure*}
	\centering
	\begin{minipage}[b]{0.997\columnwidth}
		\centering
		\includegraphics[scale=0.8]{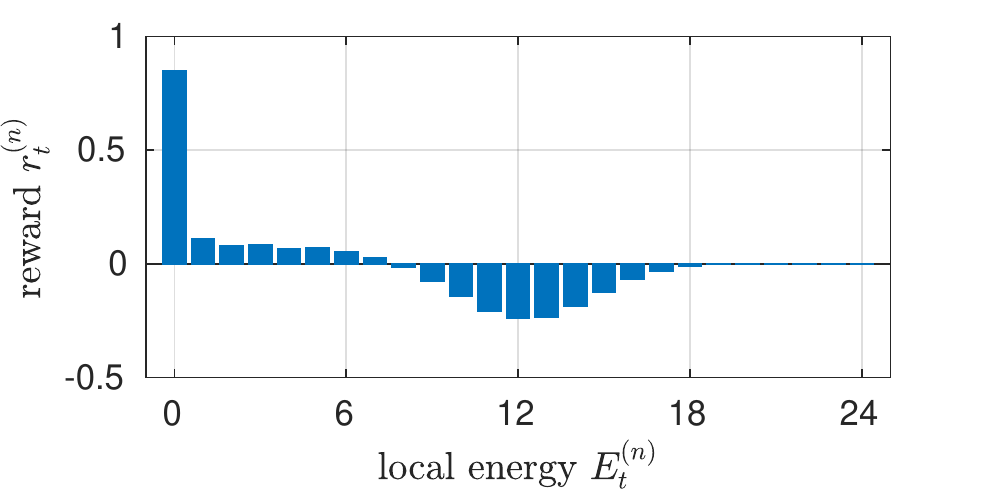}
		\caption{Learned reward function for the Ising model, averaged over 100 Monte Carlo experiments.}
		\label{fig:rewardIsing}
	\end{minipage}
	\hspace{0.75\columnsep}
	\begin{minipage}[b]{0.997\columnwidth}
		\centering
		\vspace{0.1cm}
		\includegraphics[scale=0.8]{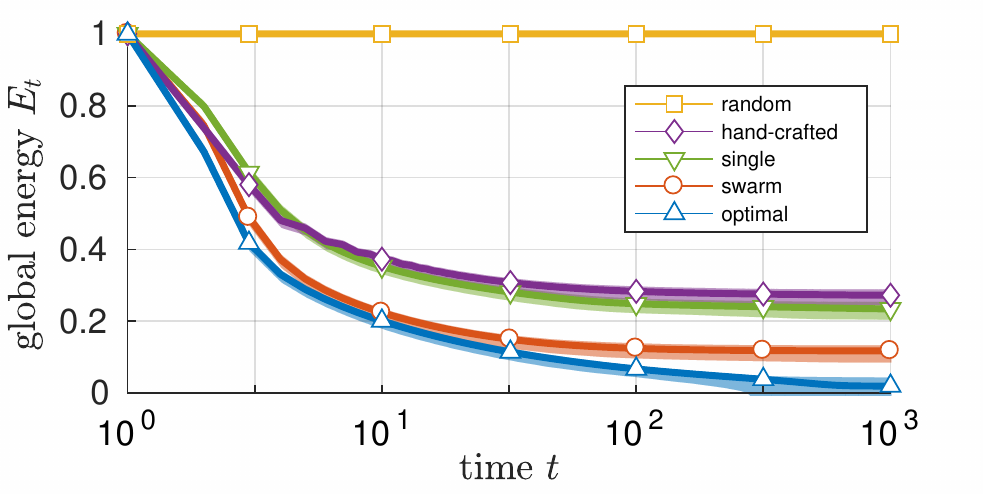}
		\caption{Slopes of the global energy $E_t$ in the Ising model. The graphs are analogous to those in \Cref{fig:orderParameter}.}
		\label{fig:energy}
	\end{minipage}
\end{figure*}

\subsection{The Ising Model}
\noindent In our second experiment, we apply the IRL framework to the well-known Ising model \cite{ising1925beitrag} which, %
in our case, consists of a finite grid of atoms (\ie\ agents) of size $100\times100$. Each agent has an individual spin $q^{(n)}_t \in \lbrace +1,-1 \rbrace$ which, together with its position on the grid, forms its local state, $s^{(n)}_t\defeq(x^{(n)},y^{(n)},q^{(n)}_t)$. For our experiment, we consider a static $5\times5$-neighborhood system, meaning that each agent %
interacts only with its 24 closest neighbors (\ie\ agents with a maximum Chebyshev distance %
of 2). 
Based on this neighborhood structure, we define the global system energy as
\begin{equation*}
	E_t \defeq \sum\limits_{n=1}^N \sum\limits_{m\in\mathcal{N}_n} %
	\indicator(q^{(n)}_t \neq q^{(m)}_t) = \sum\limits_{n=1}^N E^{(n)}_t \eqkomma
\end{equation*}
where $\mathcal{N}_n$ and $E^{(n)}_t$ are the neighborhood and local energy contribution of agent $n$, and $\indicator(\cdot)$ denotes %
the indicator function. Like the order parameter for the Vicsek model, the global energy serves as a measure for the total alignment of the system, with zero energy indicating complete state synchronization.
In our experiment, we consider two possible actions available to the agents, \ie\ \textit{keep the current spin} and \textit{flip the spin}. The system dynamics are chosen such that the agent transitions to the desired state with probability 1. As before, we give the agents the ability to monitor their local misalignment, this time provided in the form of their individual energy contributions, \ie\ $o^{(n)}_t=\xi^{(n)}(s_t)\defeq E^{(n)}_t$.

A meaningful goal for the system is to reach a global state configuration of minimum energy. Again, we are interested in learning %
a behavioral model for this task from expert trajectories. In this case, our expert system performs a local majority voting using a policy which lets the agents %
adopt the spin of the majority of their neighbors. 
Essentially, this policy implements a synchronous version of the \textit{iterated conditional modes} algorithm~\cite{besag1986statistical}, which %
is guaranteed to translate the system to a state of locally minimum energy. 

Figures \ref{fig:rewardIsing} and \ref{fig:energy} depict, respectively, the learned mean reward function %
and the slopes of the global energy %
for the different policies. As in the previous example, the extracted reward function explains the expert behavior well\footnote{\scriptsize Note that assigning a neutral reward to states of high local energy is reasonable, since a strong local misalignment indicates high synchronization of the opposite spin in the neighborhood.} and we observe the same qualitative performance improvement as %
for the Vicsek system, both when compared to the single-agent estimation scheme and to the hand-crafted model. %

\section{Conclusion \& Discussion}
\label{section:conclusion}
\noindent Our objective in this paper has been to %
extend the concept of IRL to homogeneous multi-agent systems, called swarms, in order to learn a local reward function from observed global dynamics that is able to explain the emergent behavior of a system. %
By exploiting the homogeneity of the newly introduced swarMDP model, we showed that both value estimation and policy update required for the IRL procedure can be performed based on local experience gathered at the agent level. The so-obtained reward function was provided as input to a novel learning scheme to build a local policy model which mimics the expert behavior.
We demonstrated our framework on two types of system dynamics where we achieved a performance close to that of the expert system. %

Nevertheless, there remain some open questions. In the process of IRL, we have tacitly assumed that the %
expert behavior can be reconstructed based on local interactions. Of course, this is a reasonable assumption for self-organizing systems which naturally operate in a decentralized manner. For arbitrary expert systems, however, we cannot exclude the possibility that the agents %
are instructed by a central controller which has access to the global system state. This brings us back to the following questions: When is it possible to reconstruct global behavior based on local information? If it is not possible for a given task, how well can we approximate the centralized solution by optimizing local values?

In an attempt to understand the above mentioned questions, we propose the following characterization of the reward function that would make a local policy optimal in a swarm.
To this end, we enumerate the swarm states and observations by $\mathcal{S}^N=\lbrace s_i \rbrace_{i=1}^{K}$ and $\mathcal{O}=\lbrace o_i \rbrace_{i=1}^{L}$, respectively. Furthermore, we fix an agent $n$ and define matrices $\mat{P}_o$ and $\lbrace\mat{P}_a\rbrace_{a=1}^{|\mathcal{A}|}$, where $[\mat{P}_o]_{ij} = P(s_j \given  o^{(n)}=o_i, \pi)$ and $[\mat{P}_a]_{ij} = P^{(n)}(s_j \given s_i,a, \pi)$. %
Finally, %
we represent the reward function as a vector,  \ie\ \mbox{$\mat{R}= (R(\xi^{(n)}(s_1) ), \ldots, R(\xi^{(n)}(s_{K}) )) \trans$}. %

\begin{MyProposition} %
	Consider a swarm $  ( N, \mathbb{A}, T, \xi ) $ %
	of agents $\mathbb{A} =( \mathcal{S}, \mathcal{O} , \mathcal{A}, R, \pi )$ and a discount factor $\gamma \in [0,1)$. Then, a policy $\pi:\mathcal{O}\rightarrow \mathcal{A}$ given by $\pi(o) \defeq a_1 $ is optimal\footnote{\scriptsize %
		We can ensure that $\pi(o)=a_1$ by renaming actions accordingly~\cite{ng2000algorithms}.} 
	with respect to $V(o\given\pi)$ if and only if %
	the reward $R$ satisfies
	\begin{equation}
	\mat{P}_o (\mat{P}_{a_1}- \mat{P}_a)(\mat{I}-\gamma \mat{P}_{a_1})^{-1} \mat{R} \ge 0  \quad \forall a\in\mathcal{A} \eqpunkt
	\label{eq:Characterization}
	\end{equation}
	\label{Proposition:Characterization}
\end{MyProposition}
\vspace{-2\baselineskip}
\begin{proof} %
	\label{Proof:Characterization}
	Expressing %
	\cref{eq:privateValue} using vector notation, we get 
	\begin{equation*}
		\mat{V}_{s|\pi} = (\mat{I}- \gamma \mat{P}_{a_1})^{-1}\mat{R} \eqkomma
	\end{equation*}
	where $\mat{V}_{s|\pi} = ( V^{(n)}(s_1 \given \pi),\ldots, V^{(n)}(s_{K} \given \pi))  \trans$. According to Prop.~\ref{Proposition:Limiting Value}, the corresponding limiting value function is %
	\begin{equation*}
		V(o\given\pi) =  \sum\limits_{i=1}^K P(s_i \given o^{(n)}=o,\pi) V^{(n)}(s_i\given\pi).
	\end{equation*}
	Rewritten in vector notation, we obtain
	\begin{equation}
	\mat{V}_{o|\pi} = \mat{P}_o \mat{V}_{s|\pi}  \eqkomma
	\label{eq:localVSglobal}
	\end{equation}
	where $\mat{V}_{o|\pi} = ( V(o_1\given\pi),\ldots, V(o_{L}\given\pi)) \trans$. Now, $\pi(o)=a_1 $ is optimal if and only if for all $a \in \mathcal{A}, o \in \mathcal{O}$
	\begin{align*}
		&	Q^{(n)}(o,a_1\given\pi)  \ge Q^{(n)}(o,a\given\pi)\\
		& \Leftrightarrow \sum_{i=1}^K \prob(s_i \given o^{(n)}=o, \pi) \sum_{j=1}^K \prob^{(n)}(s_j \given s_i, a_1 , \pi) V^{(n)} (s_j\given\pi)  \\
		& \phantom{\Leftrightarrow}\;\geq \sum_{i=1}^K \prob(s_i \given o^{(n)}=o, \pi) \sum_{j=1}^K \prob^{(n)}(s_j \given s_i, a, \pi) V^{(n)}(s_j\given\pi) \\
		& \Leftrightarrow \mat{P}_o (\mat{P}_{a_1} - \mat{P}_a)  \mat{V}_{s|\pi}  \ge 0 \\
		&	\Leftrightarrow \mat{P}_o (\mat{P}_{a_1} - \mat{P}_a)   (\mat{I}- \gamma \mat{P}_{a_1})^{-1} \mat{R}  \ge 0  \eqpunkt \hspace{2.8cm} \square
	\end{align*} %
	\renewcommand{\qedsymbol}{}
	\vspace{-1.3\baselineskip}
\end{proof}

\begin{myRemark} Following a similar derivation as in \cite{ng2000algorithms}, we obtain the characterization set with respect to $\mat{V}_{s|\pi}$ as
	\begin{equation}
	(\mat{P}_{a_1}- \mat{P}_a)(\mat{I}-\gamma \mat{P}_{a_1}) \mat{R} \ge 0 \eqpunkt
	\label{eq:globalCharactarization}
	\end{equation}
\end{myRemark}
\vspace{0.7\baselineskip}
\noindent Notice that, as \cref{eq:globalCharactarization} implies \cref{eq:Characterization}, an $\mat{R}$ that makes $\pi(o)$ optimal for $\mat{V}_{s|\pi}$, also makes it optimal for $\mat{V}_{o|\pi}$. Therefore, denoting by $\mathcal{R}_L$ and $\mathcal{R}_{G}$ the solution sets corresponding to the local and global values $\mat{V}_{o|\pi}$ and $\mat{V}_{s|\pi}$, we conclude
\begin{equation*}
	\mathcal{R}_G \subseteq \mathcal{R}_L \eqkomma
\end{equation*}
with equality in the trivial case where observation~$o$ is sufficient to determine the swarm state~$s$. It is therefore immediate that, as long as there is uncertainty about the swarm state, local planning can only guarantee globally optimal behavior  in an average sense as pronounced by $\mat{P}_o$ (see \cref{eq:localVSglobal}).

\section*{Acknowledgment}
W.~R.~KhudaBukhsh was supported by the German \mbox{Research} Foundation (DFG) within the Collaborative~Research Center (CRC) 1053 -- MAKI. H.~Koeppl acknowledges~the~ support of the LOEWE Research Priority Program \mbox{CompuGene}.

\bibliographystyle{abbrv}
\bibliography{bibliography}  %
\end{document}